\documentclass[runningheads]{llncs}
\usepackage{graphicx}
\usepackage{hyperref}

\usepackage{subcaption}
\usepackage[ruled,vlined,linesnumbered]{algorithm2e}
\usepackage{amsmath}
\usepackage[table,xcdraw]{xcolor}
\usepackage{multirow}
\usepackage{amssymb}

\usepackage{paralist}

\begin{document}
\title{Optimality-based Analysis of XCSF Compaction in Discrete Reinforcement Learning\thanks{The final authenticated publication is available online at \url{https://doi.org/10.1007/978-3-030-58115-2_33}}}
\titlerunning{Optimality-based Analysis of XCSF Compaction in Discrete RL}
%
\author{Jordan T. Bishop \and Marcus Gallagher}
\authorrunning{J.T. Bishop \and M. Gallagher}
%
\institute{School of Information Technology and Electrical Engineering, The University of Queensland, Brisbane, Queensland, 4072, Australia\\
\email{\{j.bishop,marcusg\}@uq.edu.au}}
\maketitle              
\vspace{-0.75em}
\begin{abstract}
Learning classifier systems (LCSs) are population-based predictive systems that were originally envisioned as agents to act in reinforcement learning (RL) environments. These systems can suffer from population bloat and so are amenable to compaction techniques that try to strike a balance between population size and performance. A well-studied LCS architecture is XCSF, which in the RL setting acts as a Q-function approximator. We apply XCSF to a deterministic and stochastic variant of the FrozenLake8x8 environment from OpenAI Gym, with its performance compared in terms of function approximation error and policy accuracy to the optimal Q-functions and policies produced by solving the environments via dynamic programming. We then introduce a novel compaction algorithm (Greedy Niche Mass Compaction --- GNMC) and study its operation on XCSF's trained populations. Results show that given a suitable parametrisation, GNMC preserves or even slightly improves function approximation error while yielding a significant reduction in population size. Reasonable preservation of policy accuracy also occurs, and we link this metric to the commonly used steps-to-goal metric in maze-like environments, illustrating how the metrics are complementary rather than competitive.

\end{abstract}
\section{Introduction}
Reinforcement learning (RL) is characterised by an agent learning a behavioural policy in an environment by means of maximising a reward signal. Learning Classifier Systems (LCSs) are a paradigm of cognitive systems that originated via representing agents in this framework, although due to flexibility in implementation have also been widely adapted to other kinds of machine learning (ML) tasks such as classification and clustering \cite{urbanowicz_introduction_2017}. The most widely-studied LCS architecture to date, Wilson's XCS \cite{wilson_classifier_1995}, is at its heart a reinforcement learner. More recently, an extension of XCS to allow for function approximation, dubbed XCSF \cite{wilson_classifiers_2001}, has been successfully used in the RL setting for value function approximation \cite{lanzi_xcs_2005,lanzi_xcs_2005-1}.

LCSs utilise a combination of evolutionary computation and ML techniques to create population-based solutions to prediction problems. The most common style of LCSs are Michigan-style LCSs, where each individual (classifier) in the population represents a partial solution, and classifiers co-operate in a potentially overlapping piecewise ensemble to define an overall solution \cite{urbanowicz_introduction_2017}. A general issue with Michigan-style LCSs is that of population bloat and/or redundancy. Since these systems learn in an online fashion and regularly refine their population via a genetic algorithm (GA), after learning is complete there are often members of the population that have not had time to properly adapt to the environment and form accurate predictions.

A common way to deal with this issue is to perform a post-processing \textit{compaction} procedure after the system is trained in order to remove low-quality classifiers from the population \cite{urbanowicz_introduction_2017}. Compaction seeks to shrink the population size as much as possible while simultaneously minimising degradation of predictive performance. This is often done as part of an analysis pipeline where the system is being used to ``mine'' knowledge from the problem via interpretation of the compacted population \cite{urbanowicz_analysis_2012}. Wilson originally described a compaction algorithm for a variant of XCS trained on a classification problem in \cite{wilson_compact_2002}, and other algorithms such as those detailed in \cite{fu_modified_2002,dixon_ruleset_2003,tan_rapid_2013} extended this line of work. These algorithms all incorporate some kind of greedy heuristic to preferentially retain some classifiers over others, and mainly use metrics related to classification performance. Since compaction is related to knowledge discovery, other works have focused more on this latter task \cite{butz_knowledge_2004,kharbat_new_2007,urbanowicz_analysis_2012}. What all these works have in common is that they study compaction in the context of supervised learning.

In this work we apply XCSF to discrete maze-like RL environments and perform compaction on the trained populations. We are interested in measuring the performance of XCSF with respect to the optimal solutions to the environments, and investigating how performance is impacted when performing compaction. As part of our analysis we introduce a novel compaction algorithm called Greedy Niche Mass Compaction (GNMC) as a generalisation of previous work. We also attempt to connect our optimality metrics to the steps-to-goal metric used by other work applying LCSs to maze-like environments.
\section{Background}
\subsection{Reinforcement Learning}
RL environments can be modelled as a Markov Decision Process (MDP), defined by components $ (S, A, T, R, \gamma) $ where $ S $ is the state space, $ A $ is the action space, $ T $ is the transition function, $ R $ is the reward function and $ \gamma \in [0, 1] $ is the reward discount factor \cite{sutton_reinforcement_2018}. We consider the case where the agent interacting with the environment seeks to learn a \textit{deterministic} behavioural policy $ \pi : S \rightarrow A $. From the agent's perspective, $ T $ and $ R$ are unknown and so learning becomes an act of balancing exploration with exploitation to sample from $ T $ and $ R $ in order to construct $ \pi $. If the full definition of the MDP is known, dynamic programming methods such as value iteration can be used to exhaustively obtain an optimal solution to the problem.

Value iteration yields an optimal Q-function $ Q^* : S \times A \rightarrow \mathbb{R} $, which maps each state-action pair $ (s, a) \in S \times A $ to a real number representing the utility of the pair: the expected amount of cumulative discounted reward that can be obtained from performing action $ a $ in state $ s $, and acting optimally thereafter. An optimal policy $ \pi^* $ can then be constructed by acting greedily with respect to $ Q^*$. One of the main approaches to RL is to have the agent build an approximation $ \hat{Q} $ to $ Q^*$ from its environmental experience using e.g. temporal difference learning techniques such as Q-learning \cite{sutton_reinforcement_2018}. The agent's approximation $ \hat{\pi}$ to $\pi^*$ can then be constructed by acting greedily with respect to $\hat{Q}$.

\subsection{XCSF}
XCSF is an LCS architecture designed to perform function approximation. It differs from XCS in that classifiers \textit{compute} their predictions as a function of their inputs, instead of predicting a scalar value. The system operates by adaptively partitioning the input space into subspaces with classifiers (evolutionary component), in tandem with forming approximations to the target function in the subspaces (ML component) \cite{butz_learning_2015}.

Classifiers take the form of $ IF\ condition\ THEN\ action $ rules. Partitioning is accomplished by specifying the \textit{rule representation} to be used by conditions. Common choices include hyperrectangles \cite{wilson_classifiers_2001,lanzi_xcs_2005} and hyperellipsoids \cite{butz_function_2008}. In the simplest case, linear functions can be used as a prediction scheme but extensions to the non-linear case have been investigated \cite{lanzi_extending_2005}.
Additionally, classifiers have a number of parameters, denoted as $cl.param$, that store or calculate information related to them; the following parameters being important in this work:
\textit{fitness} --- the predictive accuracy of a classifier relative to other classifiers in the action set(s) (defined below) that it participates in, \textit{numerosity} --- the number of copies of a classifier present in the population (necessary because the GA may produce classifiers with duplicate rules), \textit{generality} --- a quantity in the range $(0, 1]$ representing the fraction of the input space covered by the classifier's condition. Numerosity yields the concept of \textit{macroclassifiers} and \textit{microclassifiers}, defined as the classifiers in the population with unique rule structures (possibly having numerosity $ > 1$) and individual copies of these unique classifiers, respectively.

Applied as a RL method, XCSF uses a Q-learning style reinforcement component to form classifier predictions. The overall system output $\hat{Q}$ is computed for each $ (s, a) \in S \times A $ according to:
\begin{equation}
 \hat{Q}(s, a) = \frac{\sum_{cl \in [A]}cl.prediction(s)\,\cdot\,cl.fitness}{\sum_{cl \in [A]}cl.fitness}
\label{eqn:pa-calc}
\end{equation}
$[A]$ is termed an \textit{action set} and contains classifiers in the population whose conditions match $ s $ and who advocate action $a$, i.e. XCSF's current knowledge about a particular \textit{niche} of the environment.
\section{Environments}
\label{sec:envs}
\begin{figure}
    \begin{subfigure}{.475\textwidth}
        \centering
        \includegraphics[width=.7\linewidth]{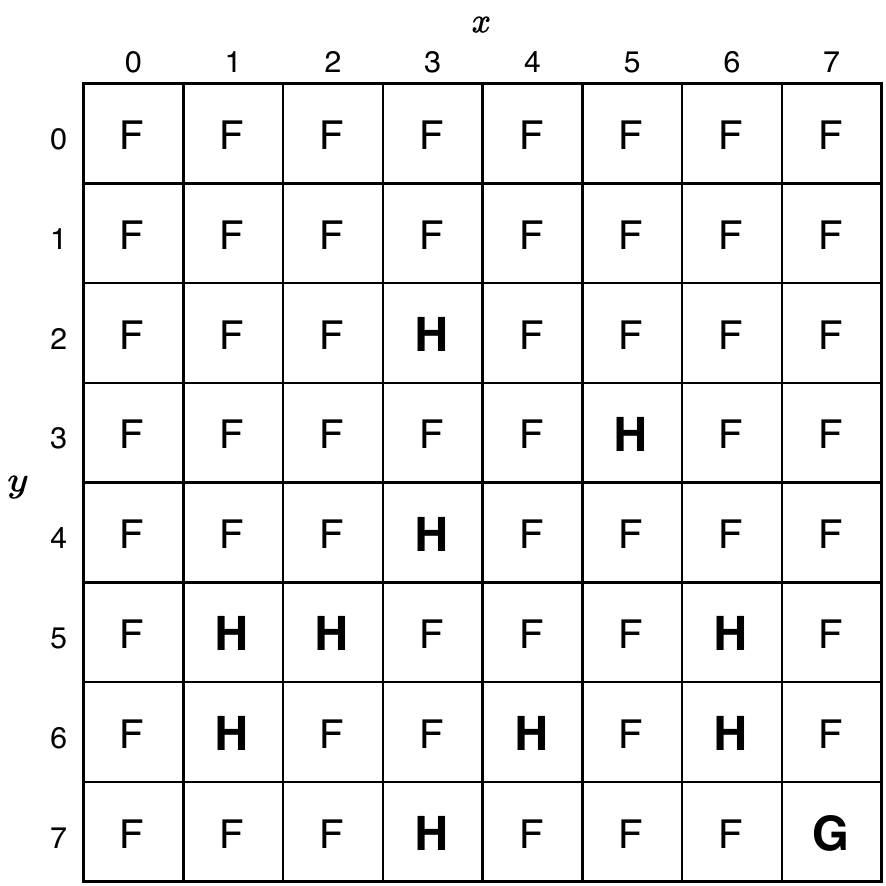}
        \caption{Environment grid structure. Cell labels: F = frozen, H = hole, G = goal.}
        \label{subfig:frozenlake-grid}
    \end{subfigure}\hspace{.025\textwidth}
    \begin{subfigure}{.475\textwidth}
        \centering
        \includegraphics[width=.6\linewidth]{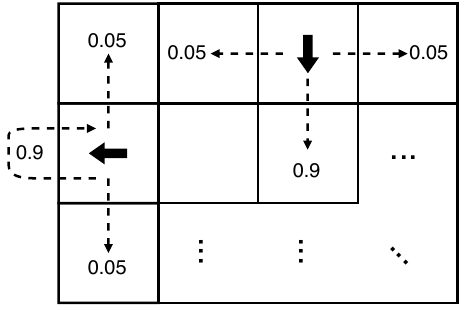}
        \caption{Transition dynamics with $p_{slip}=0.1$.
         With probability 0.1 the agent slips to one of the directions perpendicular to its intended direction (mass shared equally across both possible slip directions). As illustrated by the leftmost example, the agent is unable to maneuver off the grid.}
        \label{subfig:frozenlake-transitions}
    \end{subfigure}
    \caption{FrozenLake8x8 (a) structure and (b) example transition dynamics.}
\end{figure}

We consider two variations of the FrozenLake environment with grid size 8 (FrozenLake8x8) from OpenAI Gym\footnote{\url{https://gym.openai.com/envs/FrozenLake8x8-v0/}}. FrozenLake is an episodic, fully observable grid navigation environment. In this environment, the agent must navigate across frozen cells to reach a goal, without falling into any holes. If the agent falls into a hole the episode terminates. The state representation used is an $(x, y)$ co-ordinate representing the location of the agent in the grid, as shown in Figure~\ref{subfig:frozenlake-grid}. We use $ S $ to indicate the set of \textit{non-terminal} states (frozen cells), and $ S^T $ to represent the set of \textit{terminal} states (holes and the goal). The action space $ A = \{Left,\ Down,\ Right,\ Up\}$, constant over all $ s \in S $. 

A parameter $p_{slip}$ controls the level of stochasticity in the environmental transition dynamics. Figure~\ref{subfig:frozenlake-transitions} gives examples of transition dynamics with $ p_{slip} = 0.1 $. Transition stochasticity is global over all $ s \in S $. By default $p_{slip} = \frac{2}{3}$, which is quite high. For our variants, we consider the cases where $p_{slip} = 0$ and $ p_{slip} = 0.1$, the latter because we wish to preserve the spirit of the default case while making the problem substantially easier through lowering the amount of noise incurred in transitions and therefore the reward signal.
The reward function operates as follows: +1 if the agent transitions into G, 0 otherwise.  We set $ \gamma = 0.95 $ to ensure that there is time pressure to reach the goal. 
Note that a time step is counted even if the agent does not move to a new state after performing an action (as occurs with 90\% probability in the leftmost example of Figure~\ref{subfig:frozenlake-transitions}).

\begin{figure}
\begin{subfigure}{.475\textwidth}
  \centering
    \includegraphics[width=.7\linewidth]{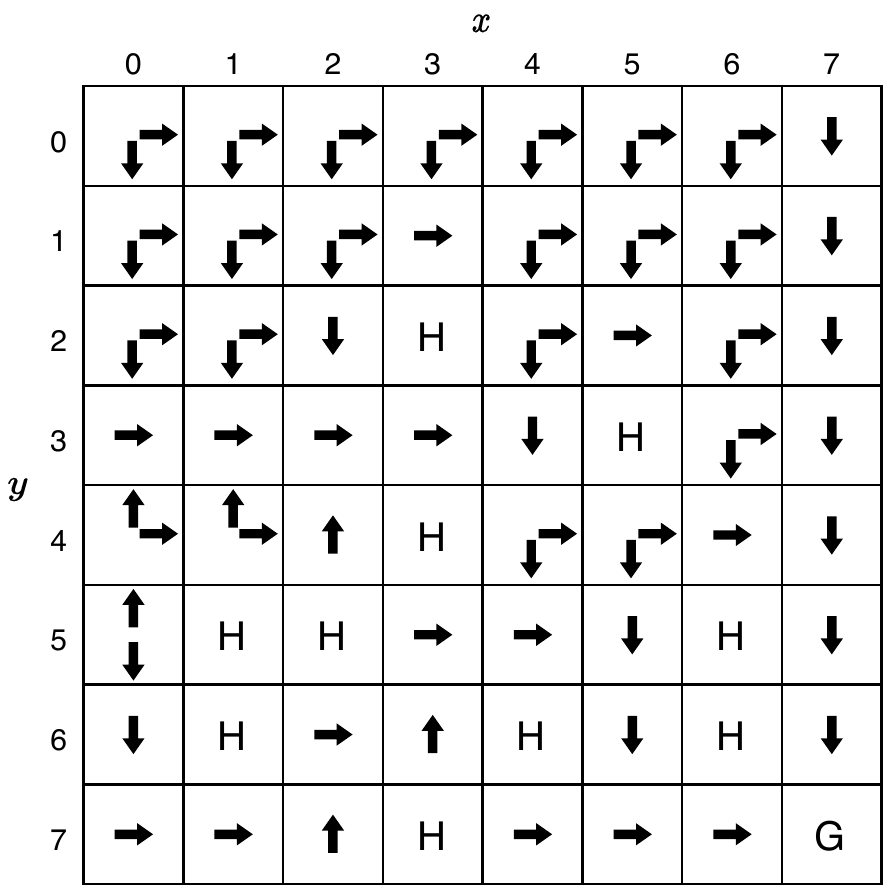}  
  \caption{$p_{slip} = 0$}
  \label{subfig:pi_star_pslip_0}
\end{subfigure}
\begin{subfigure}{.475\textwidth}
  \centering
  \includegraphics[width=.7\linewidth]{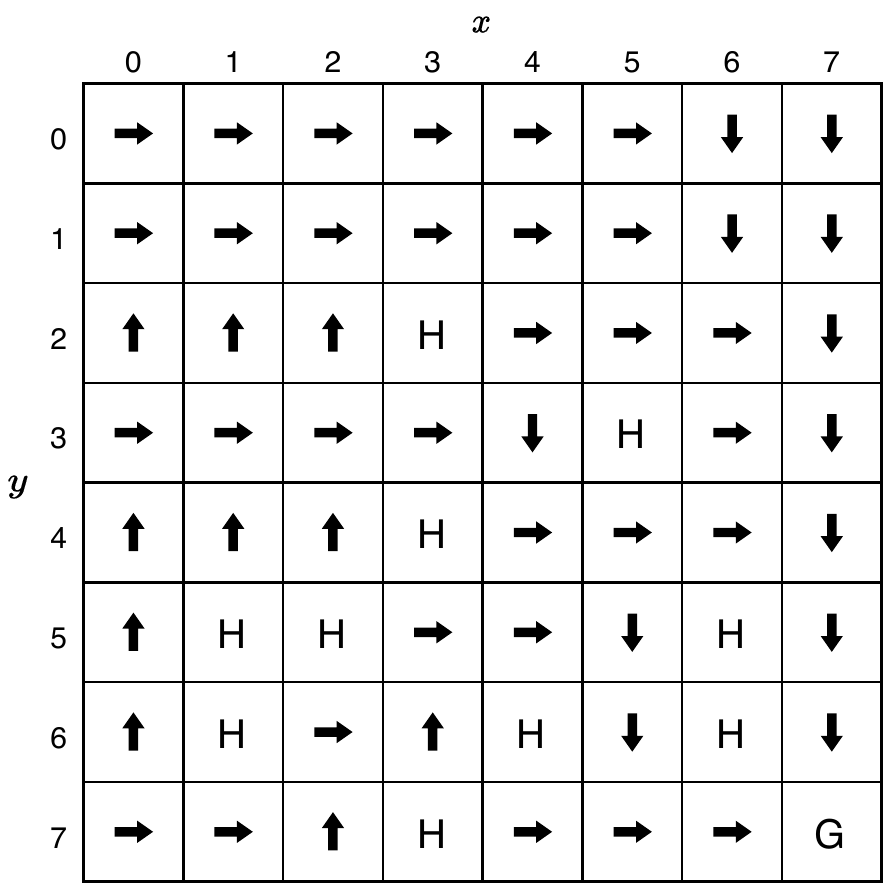}  
  \caption{$p_{slip} = 0.1$}
  \label{subfig:pi_star_pslip_0.1}
\end{subfigure}
\caption{Optimal policies for FrozenLake8x8, $ \gamma = 0.95 $.}
\label{fig:optimal-solns}
\end{figure}

Figure~\ref{fig:optimal-solns} shows the optimal policies for our two variants. In the deterministic case reaching the goal is a shortest path problem, hence in some states there are multiple optimal actions. In the stochastic case the optimal policy is more strict as there is only a single optimal action in every state.

\section{XCSF Configuration}
\label{sec:xcsf-config}
We use our own implementation of XCSF written in Python\footnote{\url{https://github.com/jtbish/piecewise}, see also \url{https://github.com/jtbish/ppsn2020} for experimental code that uses this.}, faithful to the base description of XCS given in \cite{butz_algorithmic_2002}. We use the same linear prediction scheme as in \cite{wilson_classifiers_2001}, where each classifier has an associated weight vector and its prediction is computed as a dot product between its weight vector and the input vector, and classifier weight vectors are updated via a normalised least mean squares procedure with the prediction target calculated via the system's reinforcement component. Also incorporated is the extension to XCS from \cite{lanzi_extension_1999}, termed XCS$\mu$, which is used to estimate uncertainty introduced by stochasticity in the environment. This involves adding a parameter $ \mu $ to each classifier which tracks minimum prediction error in the action sets the classifier participates in, adjusted by a separate learning rate $\beta_{\epsilon}$. 

The rule representation used is an interval-based representation, specifically an integer-valued variant of min-percentage representation \cite{dam_be_2005}.
Interval minimum alleles are retained but percentage-to-maximum alleles are replaced by ``span-to-maximum'' alleles; interval maximums calculated as: $\mathtt{max} = \mathtt{min} + \mathtt{span}$. The covering and mutation operators from \cite{dam_be_2005} are adopted and modified to work with integer values, resembling those in \cite{wilson_mining_2000}. Subsumption and calculation of condition generality are the same as in \cite{wilson_mining_2000}. GA selection is done via tournament selection and uniform crossover is applied on allele sequences.

The chosen rule representation and prediction scheme yield a system that learns linear predictions of value over rectangular regions of the input space. This is suitable for both FrozenLake8x8 environments because it exploits the fact that Q-values decay smoothly (due to discounting) when moving away from the goal. In areas of the state space where there are no holes, accurate generalisation over large areas is possible (refer to e.g. top two rows in Figure~\ref{subfig:frozenlake-grid}) so only a few classifiers are required to cover such an area. The opposite is true for areas near holes. Compared to other Q-function approximators used in RL (e.g. neural networks), XCSF has the advantage of presenting its knowledge in a piecewise, easily interpretable format that can reduced to a compact set of classifiers (as is the theme of this work).
\section{Training Experiments}
\label{sec:train}
\subsection{Setup}
\label{subsec:train-setup}
We train our implementation of XCSF described in Section~\ref{sec:xcsf-config} on the two environments detailed in Section~\ref{sec:envs}. For the first environment ($ p_{slip}=0$), the training budget is 400,000 time steps (environmental transitions) and for the second environment ($p_{slip}\,=\,0.1$) the budget is doubled to 800,000 time steps.
Hyperparameters for both cases are:
$N{=}5000$, $\beta{=}0.1$, $\beta_{\epsilon}{=}0.05$, $\alpha{=}0.1$, $\epsilon_0{=}0.01$, $\nu{=}5$, $\gamma{=}0.95$, $\theta_{GA}{=}50$, $\tau{=}0.5$, $\chi{=}1.0$, $\upsilon{=}0.5$, $\mu{=}0.05$, $\theta_{del}{=}50$, $\delta{=}0.1$, $\theta_{sub}{=}50$, $\epsilon_I{=}10^{-3}$, $f_I{=}10^{-3}$, $\theta_{mna}{=}4$, $doGASubsumption{=}True$, \\$doActionSetSubsumption{=}False$, $r_0{=}4$, $m_0{=}4$, $x_0{=}10$, $\eta{=}0.1$.
Hyperparameter meanings correspond to those given in \cite{butz_algorithmic_2002,wilson_mining_2000,wilson_classifiers_2001,butz_learning_2015,lanzi_extension_1999}, except for $ \upsilon $ which is our addition and controls the probability of an allele being crossed over during uniform crossover. The two most critical hyperparameters are $ N $ (maximum population size in number of microclassifiers) and $\epsilon_0$ (target absolute approximation error). We tuned their values manually, along with the training budget. For other hyperparameters, we followed guidance from \cite{urbanowicz_introduction_2017,butz_algorithmic_2002,lanzi_xcs_2005}.

By default, the agent starts each episode in state $(0, 0)$, which puts a heavy emphasis on exploration to reach the goal, making learning relatively difficult. To make learning easier, we allow the agent to start a training episode in any $ s \in S $, selected uniformly at random. We can therefore safely adopt the alternating explore-exploit action selection strategy used elsewhere in the literature, i.e. $\epsilon$-greedy with a fixed value of $\epsilon\,=\,0.5$.

\subsection{Metrics}
\label{subsec:train-metrics}
Before training, we use value iteration to compute $ Q^* $ and consequently $\pi^*$ for each environment. XCSF's $ \hat{Q} $ mean absolute error (MAE) can then be calculated as:
\begin{equation}
    \frac{1}{|S||A|}\sum_{s \in S}\sum_{a \in A}|Q^*(s,a) - \hat{Q}(s,a)|
\end{equation}
MAE is used because we wish to directly compare with $\epsilon_0$. To allow for comparison between $\pi^*$ and $\hat{\pi}$, policies are encoded as a series of \textit{binary action advocacy vectors}, one for each $ s \in S$, whereby if policy $ \pi $ advocates action $ a_i $ in state $ s $, bit $ i $ of $ \pi(s) $ is set to 1, 0 otherwise. For example, following Figure~\ref{subfig:pi_star_pslip_0} the optimal actions in state $ (0, 0) $ are $\{Down,\ Right\}$. Assuming the ordering of actions is $\{Left,\ Down,\ Right,\ Up\}$, then the encoding $\pi^*\big((0, 0)\big)\,=\,[0, 1, 1, 0] $.
XCSF's $\hat{\pi}$ accuracy can then be calculated as:
\begin{equation}
    \frac{1}{|S|}\sum_{s \in S} C\big(\pi^*(s), \hat{\pi}(s)\big)
\end{equation}
where $ C $ is a Boolean function that accepts two binary action advocacy vectors, $ a^* $ and $\hat{a}$, and determines if \textit{at least one} of the actions advocated in $a^*$ is also advocated in $\hat{a}$ , i.e. determines if $ \hat{a} $ is ``correct'':
\begin{equation}
    C\big(a^*, \hat{a}\big)\,=\,\begin{cases}
        1 & \mathtt{count\_ones}\big(a^*\ \mathtt{AND}\ \hat{a}\big) > 0  \\
        0 & otherwise
    \end{cases}
\end{equation}
Reusing the previous example, if $ s\,=\,(0, 0) $, $\pi^*(s)\,=\,[0, 1, 1, 0] $ and also $\hat{\pi}(s)= [0, 0, 1, 0] $ then $ C $ returns 1 because $\hat{\pi}$ advocates one of the optimal actions, $ Right $.

\subsection{Results}
\label{subsec:train-res}
\begin{figure}
    \begin{subfigure}{.5\textwidth}
        \centering
        \includegraphics[width=\linewidth]{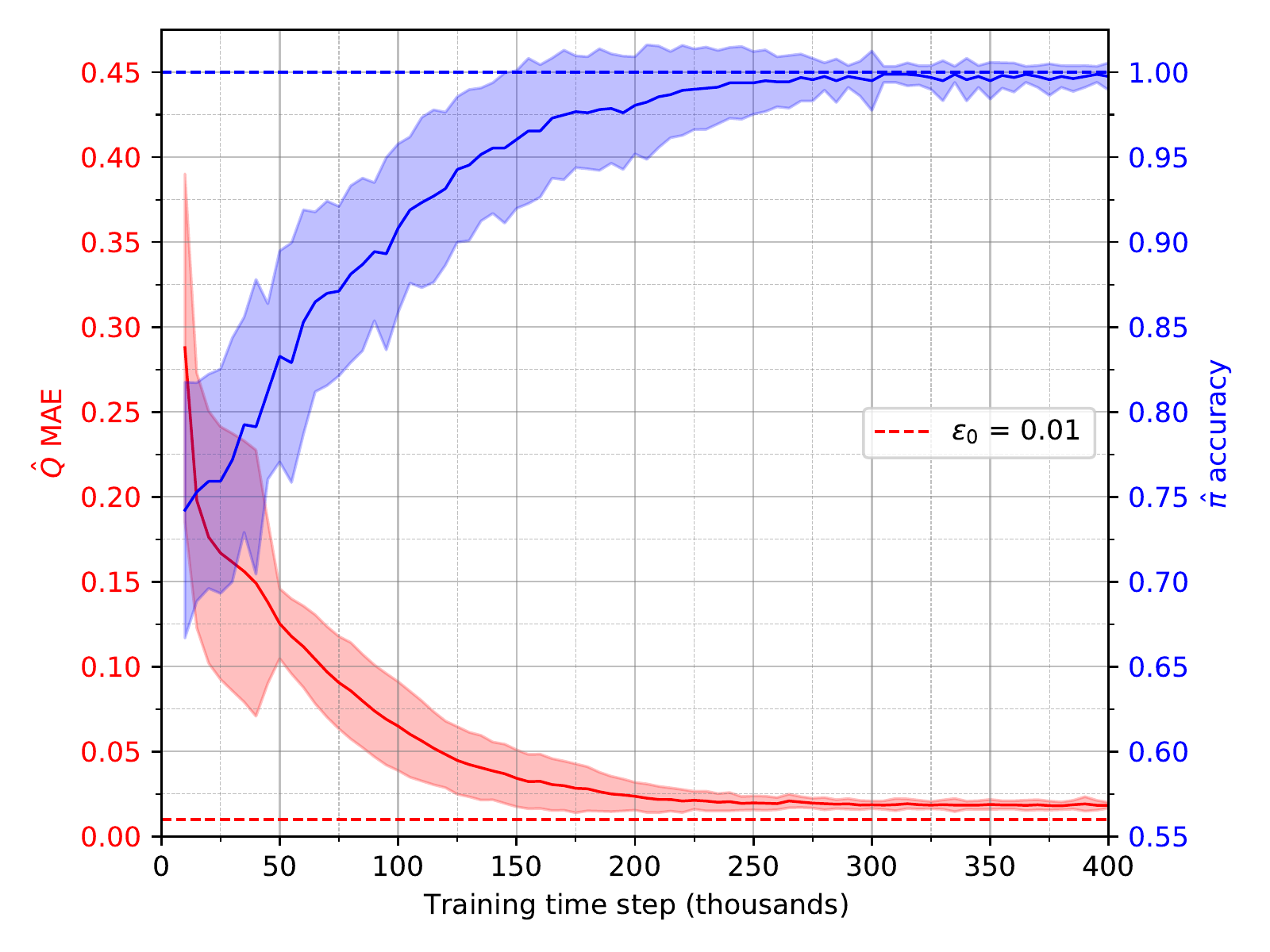}
        \caption{$p_{slip}\,=\,0$}
        \label{subfig:train-curve-pslip-0}
    \end{subfigure}
    \begin{subfigure}{.5\textwidth}
        \centering
        \includegraphics[width=\linewidth]{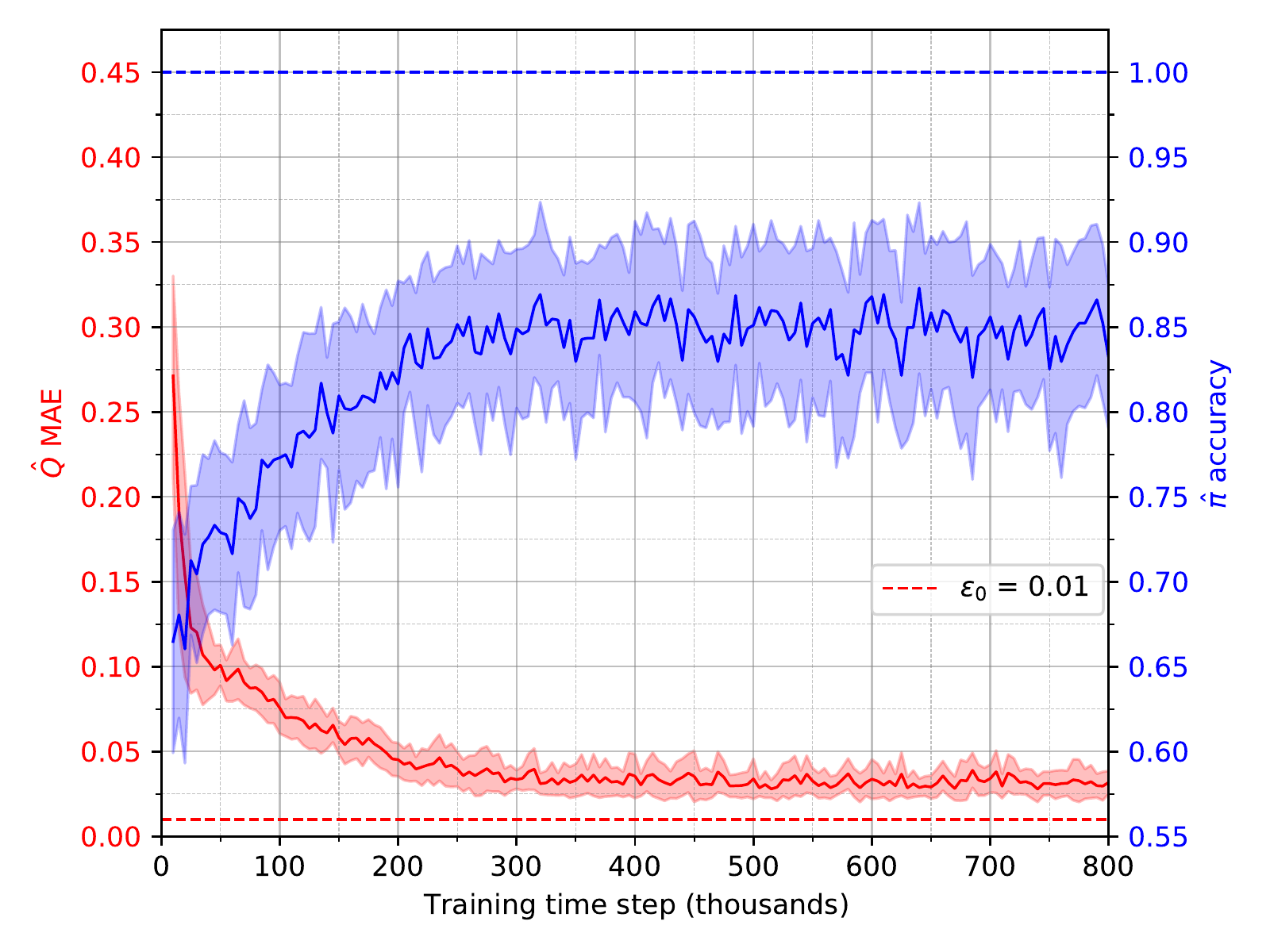}
        \caption{$p_{slip}\,=\,0.1$}
        \label{subfig:train-curve-pslip-0.1}
    \end{subfigure}
    \caption{XCSF training performance curves on FrozenLake8x8 environments. Solid lines are the mean of 30 trials, shaded regions are one standard deviation.}
    \label{fig:train-curves}
\end{figure}

Figure~\ref{fig:train-curves} shows XCSF training performance curves for both environments, measured over time are $ \hat{Q}$ MAE and $\hat{\pi}$ accuracy. In the deterministic case, XCSF converges to a small MAE that is slightly larger than the target error threshold $\epsilon_0$, with $\hat{\pi} $ accuracy very close to the maximum value of 1.
In the stochastic case, MAE is still quite small but noticeably larger than in the deterministic case, also with larger variance. $\hat{\pi}$ accuracy is significantly lower and with much larger variance.
We now investigate this reduction of $\hat{\pi}$ accuracy in the stochastic case in more detail.
Figure~\ref{fig:end-of-training-pi-correct-heatmap-pslip-0.1} shows the frequency of optimal action predictions for each $ s \in S $ over the 30 trained instances. From this we can see that XCSF is quite often predicting the optimal action in a majority of states. However, there are a few states that are degrading policy accuracy more than others. Table~\ref{tab:end-of-training-bad-states-actions} shows the distributions of actions predicted in the four states with lowest optimal action prediction frequencies, and indicates that for these states if the predicted action is not optimal ($ Up $ or $ Down $) it is at least sensible ($ Right $). Thus the situation is not as poor as it first seems.

\begin{minipage}{\textwidth}
  \begin{minipage}{0.45\textwidth}
    \centering
        \includegraphics[width=.9\linewidth]{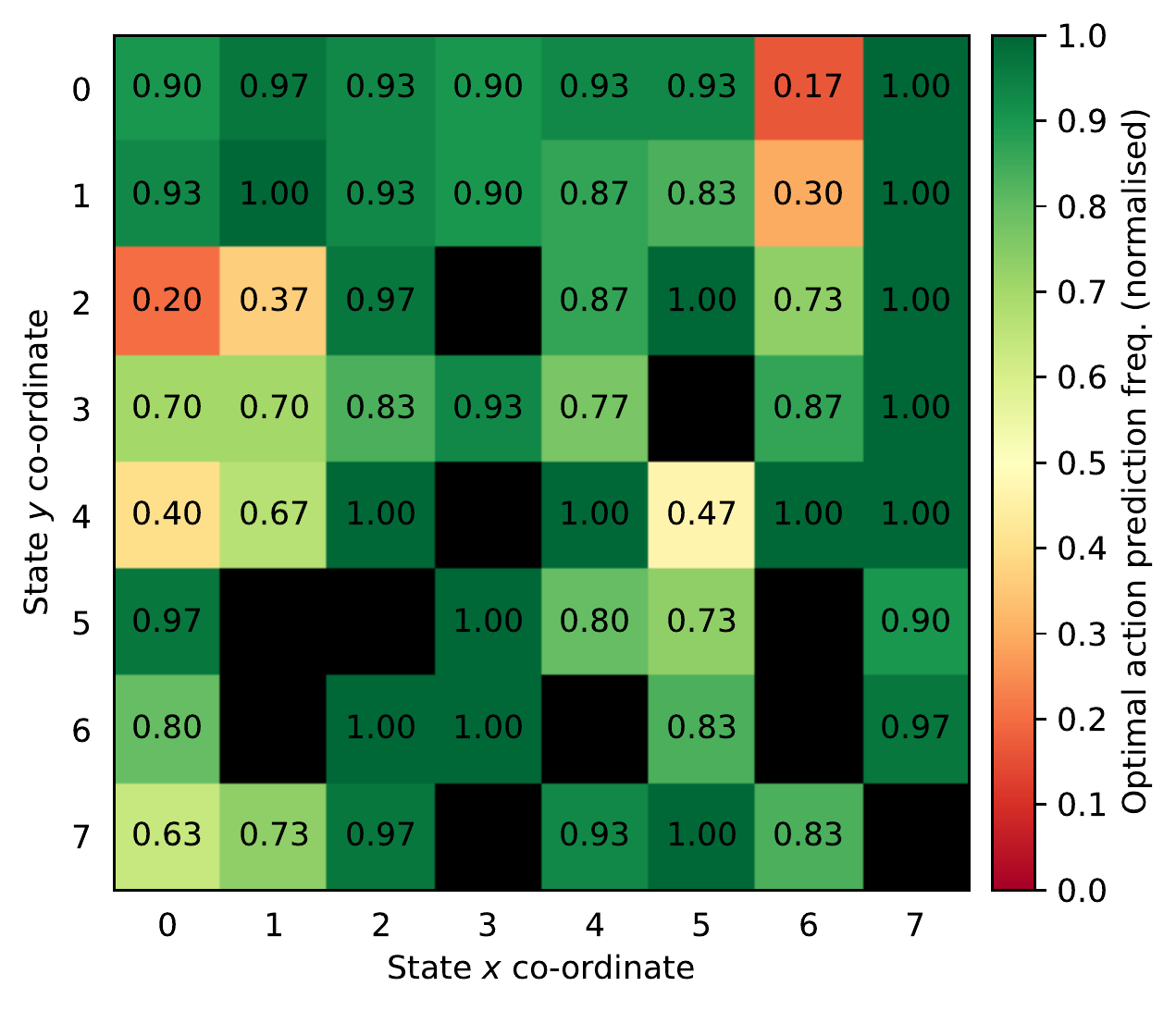}
        \captionof{figure}{XCSF optimal action prediction frequency for FrozenLake8x8 $p_{slip}\,=\,0.1$, calculated over 30 instances.}
        \label{fig:end-of-training-pi-correct-heatmap-pslip-0.1}
  \end{minipage}
  \hspace{.015\textwidth}
  \begin{minipage}{0.45\textwidth}
    \centering
    \captionof{table}{Distribution of action predictions for the four lowest frequency states in Figure~\ref{fig:end-of-training-pi-correct-heatmap-pslip-0.1}. Optimal actions for each state are set in bold.}
    \begin{tabular}{cl|l|l|l|l|l}
    \cline{3-6}
    \multicolumn{1}{l}{}                         &          & \multicolumn{4}{c|}{Action}       &                                             \\ \cline{3-7} 
    \multicolumn{1}{l}{} &
       &
      \multicolumn{1}{p{.4cm}|}{L } &
      \multicolumn{1}{p{.4cm}|}{D } &
      \multicolumn{1}{p{.4cm}|}{R } &
      \multicolumn{1}{p{.4cm}|}{U } &
      \multicolumn{1}{c|}{\begin{tabular}[c]{@{}c@{}}Optimal\\ Freq.\end{tabular}} \\ \hline
    \multicolumn{1}{|c|}{\multirow{4}{*}{\rotatebox[origin=c]{90}{State}}} & $(0, 2)$ & 0 & 0          & 24 & \textbf{6}  & \multicolumn{1}{l|}{$6/30{=}0.2$}   \\ \cline{2-7} 
    \multicolumn{1}{|c|}{}                       & $(1, 2)$ & 0 & 0          & 19 & \textbf{11} & \multicolumn{1}{l|}{$11/30{=}0.37$} \\ \cline{2-7} 
    \multicolumn{1}{|c|}{}                       & $(6, 0)$ & 0 & \textbf{5} & 25 & 0           & \multicolumn{1}{l|}{$5/30{=}0.17$}  \\ \cline{2-7} 
    \multicolumn{1}{|c|}{}                       & $(6, 1)$ & 0 & \textbf{9} & 21 & 0           & \multicolumn{1}{l|}{$9/30{=}0.3$}   \\ \hline
    \end{tabular}
    \label{tab:end-of-training-bad-states-actions}
    \end{minipage}
  \end{minipage}

\section{Compaction}
\label{sec:compaction}
We now turn to the main consideration of this work: compaction of trained XCSF populations. Algorithm~\ref{alg:gnmc} presents Greedy Niche Mass Compaction (GNMC), a novel compaction algorithm designed for use on LCS populations applied to RL environments with discrete state-action spaces.
GNMC considers all environmental action sets (niches) and greedily keeps some number of the best quality classifiers in each. The notion of ``best quality'' is defined by the parameter $ \lambda $, which is a function that assigns each classifier a mass (quality weighting) in the action set. $ \rho $ acts as a compression factor and controls the number of classifiers kept in each action set; higher values result in more classifiers being discarded.
\begin{algorithm}
    \SetAlgoLined
    \KwIn{Classifier mass function $ \lambda $, mass removal factor $ \rho \in [0, 1)$\;}
    $toKeep$ = $\emptyset$\;
    \For{$ (s, a) \in S \times A $}{\label{alg:gnmc-2}
        Create action set $[A]$ for $ (s, a) $\;
        Create set $[A]'$ by sorting $[A]$ in descending order according to $\lambda$\;
        $totalMass$ = $\sum_{cl \in [A]'} \lambda(cl)$\;
        $targetMass$ = $ (1 - \rho) \cdot totalMass $\;
        $currentMass$ = 0\;
        \While{$ currentMass < targetMass$}{
            $cl$ = next classifier in $ [A]' $\;
            $toKeep$ = $toKeep \cup \{ cl \}$\;\label{alg:gnmc-10}
            $currentMass$ += $\lambda(cl)$\;
        }
    }
    Remove classifiers not in $ toKeep $ from the population $[P]$\;
    \caption{Greedy Niche Mass Compaction (GNMC)}
    \label{alg:gnmc}
\end{algorithm}

GNMC exhibits a number of desirable properties:
\begin{compactenum}
    \item The exact number of classifiers discarded in each action set is dependent on the distribution of classifier mass; $ \rho $ is sensitive to this distribution.\label{pt1}
    \item $ \rho $ can be adjusted in a smooth manner without needing prior information about the size of action sets.
    \item It is guaranteed that no ``gaps'' in the predictive mapping are introduced, due to all action sets being considered and at least a single classifier being kept in each action set (because $\rho$ cannot equal 1).\label{pt3}
    \item Any classifiers that only match a state $ s \in S^T $ (and so have zero experience and do not contribute to overall predictions) are implicitly removed from the population because they are never added into the $ toKeep $ set; the for loop on line~\ref{alg:gnmc-2} operates only over $ S $. This occurs even when $\rho = 0$.
\end{compactenum}
Notably, simple compaction strategies such as removing all classifiers with experience less than some threshold do not uphold point~\ref{pt3} listed above. This property is crucial for function approximation in RL where a complete mapping of the state-action space is necessary.

GNMC can be viewed as a generalisation of previous work in the literature. In particular, we consider the work of Tan et al. in \cite{tan_rapid_2013}, where the authors define a compaction algorithm in the context of a classification task, called Parameter Driven Rule Compaction (PDRC).
PDRC operates by forming a correct set $ [C] $ for each environmental input (each training set data point) then keeping the classifier in $ [C] $ with the largest product of accuracy, numerosity, and generality. All other classifiers in $ [C] $ are discarded. Translating between classification and RL, $ [C] $ is analogous to $ [A] $ and classifier accuracy is analogous to fitness because Tan et al. employ a UCS (sUpervised Classifier System \cite{bernado-mansilla_accuracy-based_2003}) variant where accuracy is \textit{equivalent to} fitness.
GNMC is therefore equivalent to PDRC when the mass function $\lambda(cl) = cl.fitness \times cl.numerosity \times cl.generality$ and the mass removal factor $ \rho $ is sufficiently high so as to keep only a single classifier from each action set.

\begin{figure}[t]
\centering
\begin{subfigure}{.475\textwidth}
  \centering
  \includegraphics[width=.975\linewidth]{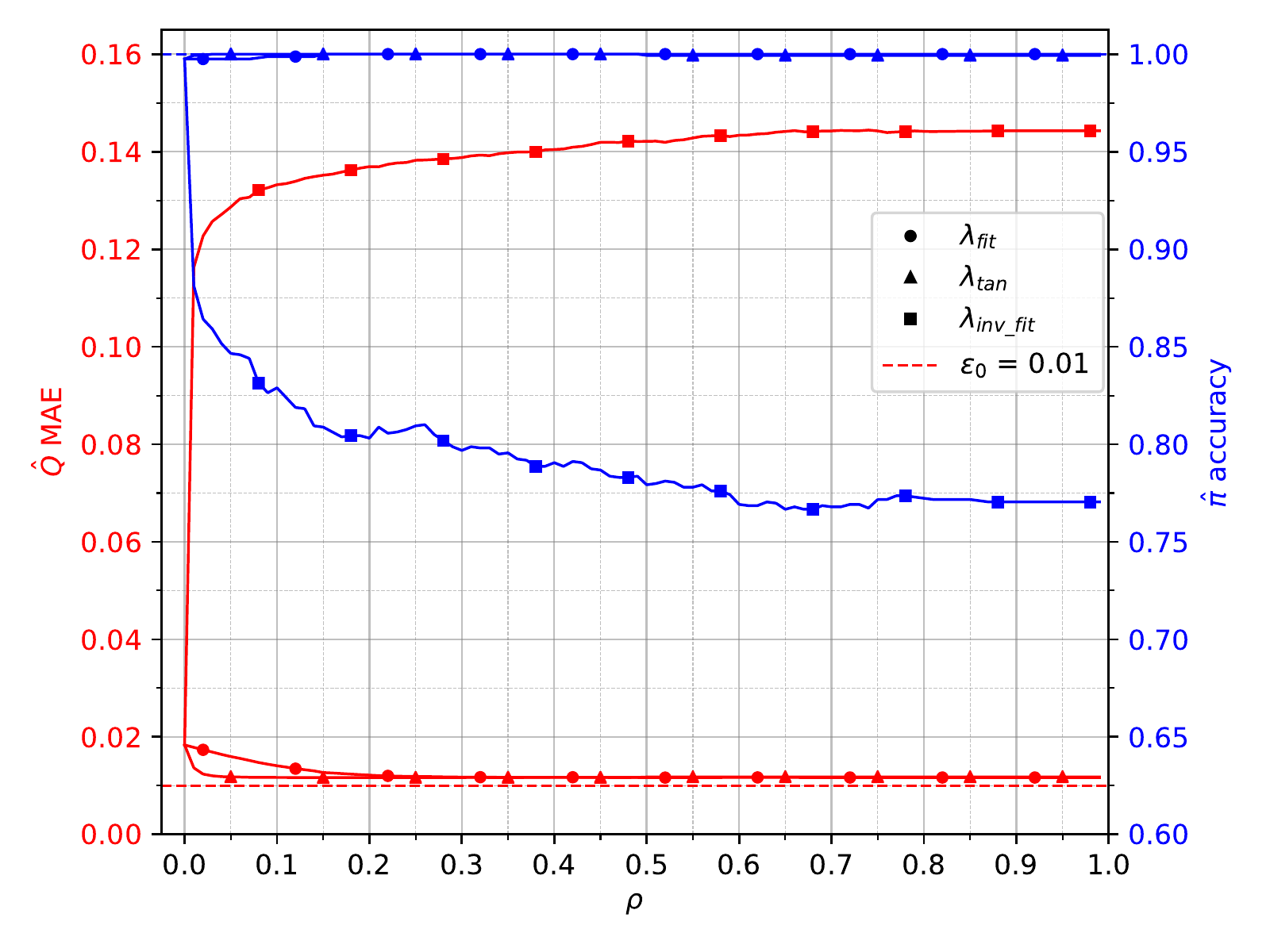}
  \caption{Performance, $ p_{slip} = 0 $}
  \label{subfig:gnmc-perf-pslip-0}
\end{subfigure}
\begin{subfigure}{.475\textwidth}
  \centering
  \includegraphics[width=.975\linewidth]{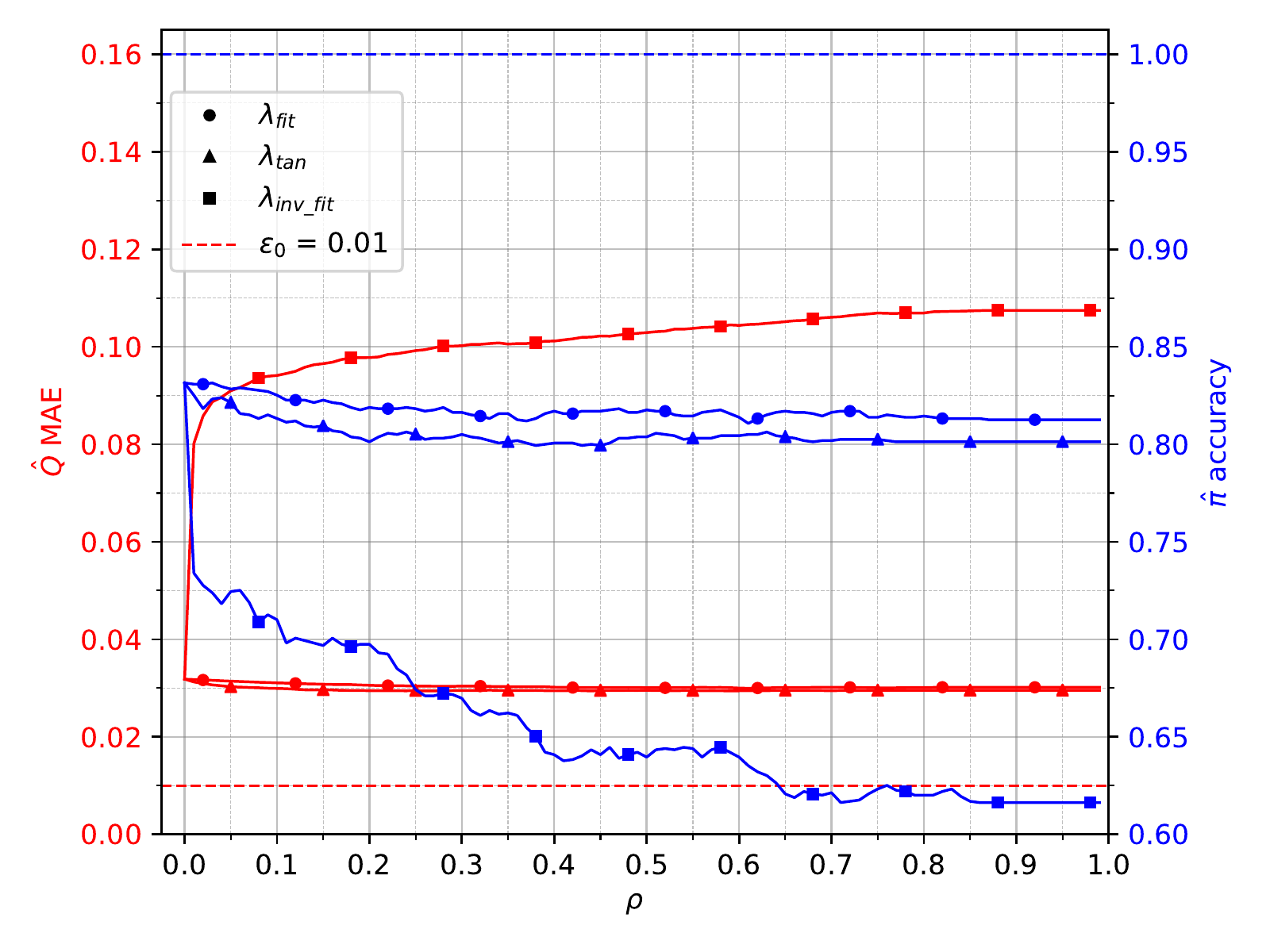}
  \caption{Performance, $ p_{slip} = 0.1 $}
  \label{subfig:gnmc-perf-pslip-0.1}
\end{subfigure}
\newline
\begin{subfigure}{.475\textwidth}
  \centering
  \includegraphics[width=.975\linewidth]{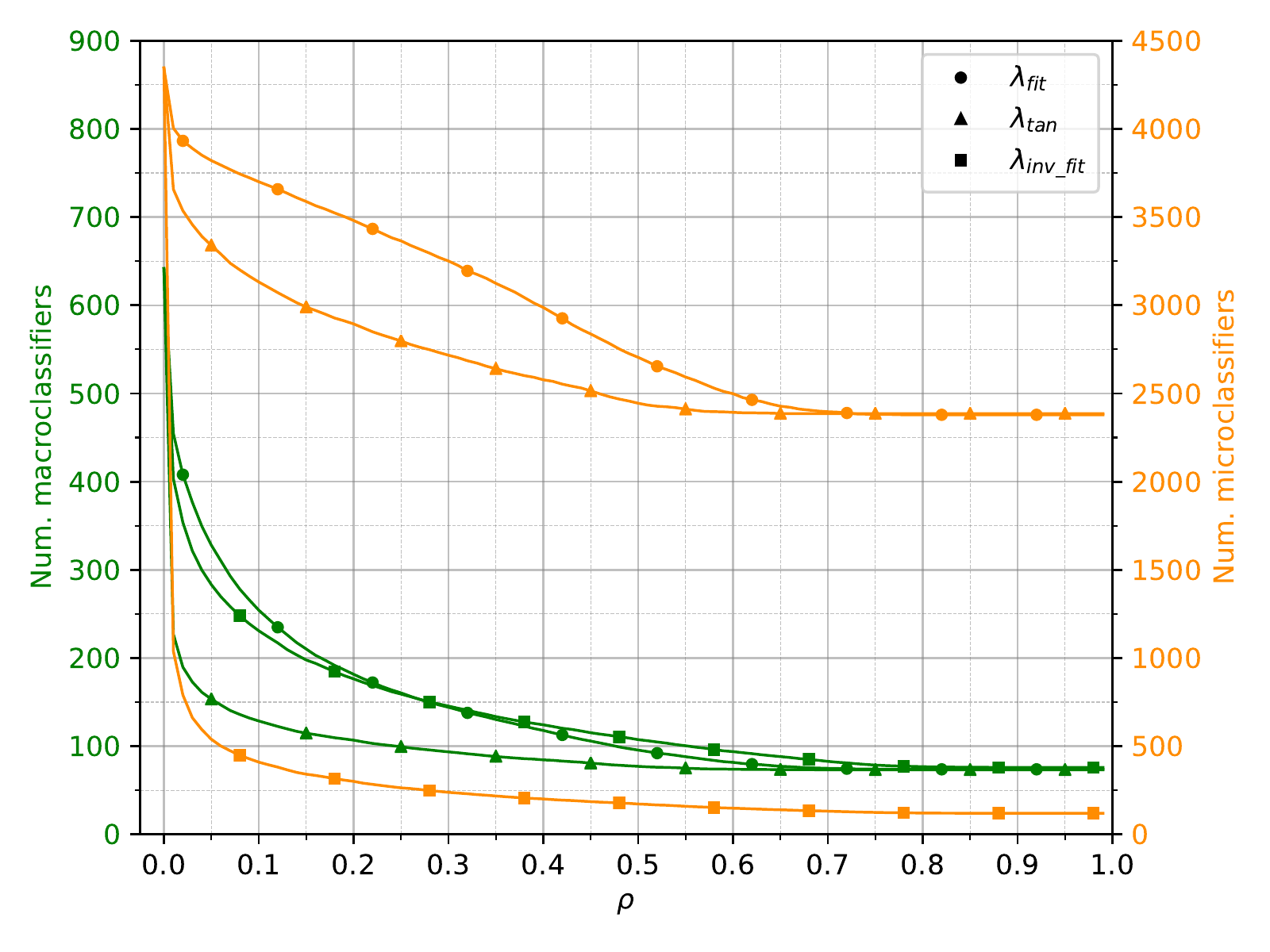}  
  \caption{Pop. size, $ p_{slip} = 0 $}
  \label{subfig:gnmc-pop-size-pslip-0}
\end{subfigure}
\begin{subfigure}{.475\textwidth}
  \centering
  \includegraphics[width=.975\linewidth]{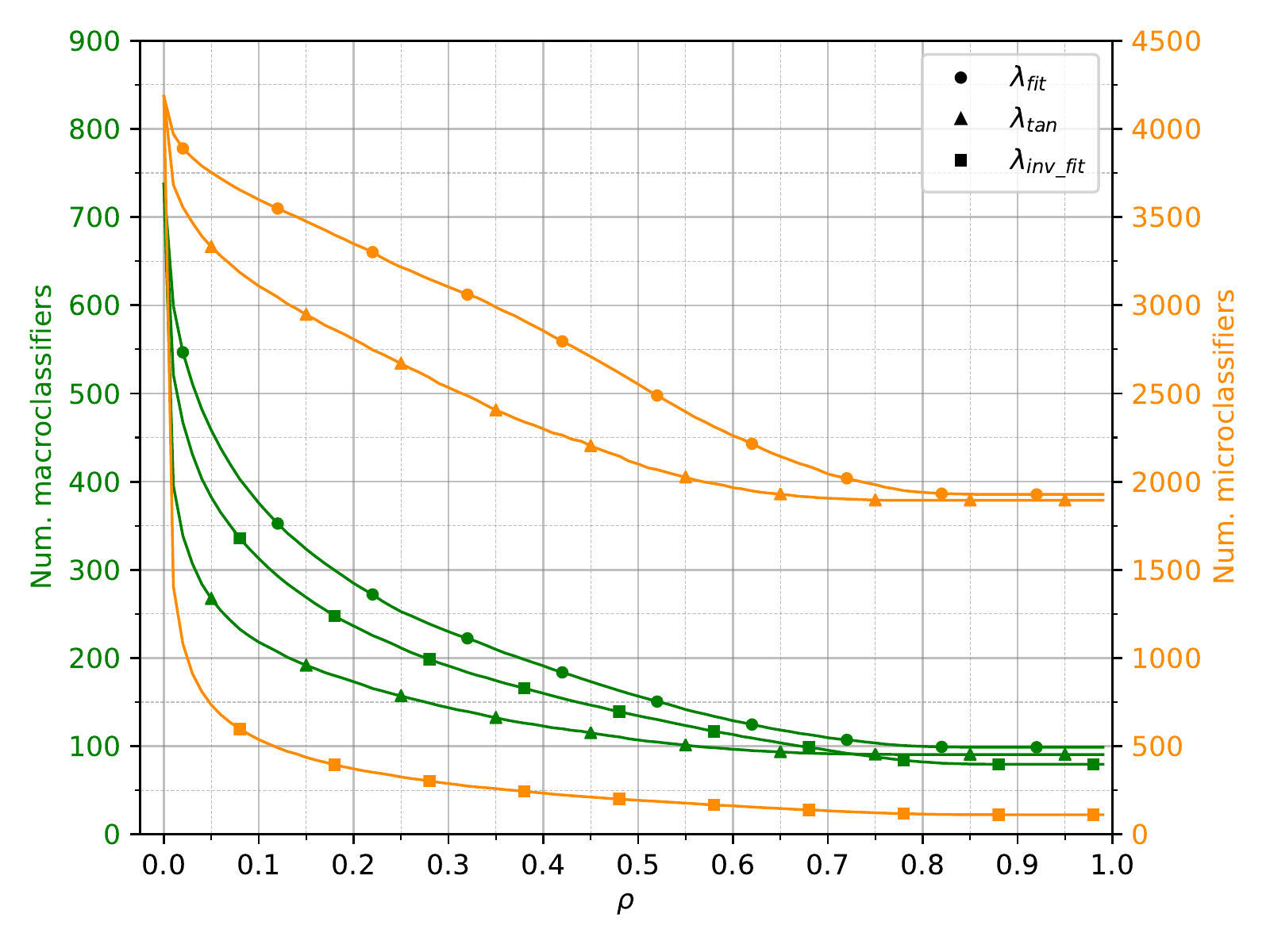}  
  \caption{Pop. size, $ p_{slip} = 0.1 $}
  \label{subfig:gnmc-pop-size-pslip-0.1}
\end{subfigure}
\caption{Results of applying GNMC with three different mass functions to both FrozenLake8x8 environments. All curves are the mean over 30 instances. Values of $ \rho $ used are from 0 to 0.99 in increments of 0.01.}
\label{fig:gnmc-res}
\end{figure}

We apply GNMC to our trained XCSF populations, considering three different mass functions, named with subscripts. The first is $\lambda_{fit} = cl.fitness $, motivated by the manner in which XCSF calculates its overall predictions: see Eqn.~\ref{eqn:pa-calc}. Classifiers with higher fitness have more weight in the overall prediction, so using fitness as a mass function makes sense. The second mass function is $\lambda_{tan} = cl.fitness \times cl.numerosity \times cl.generality $. The final mass function is an antagonistic variant of the first mass function that is designed to see what happens when GNMC is operating with ``bad information'': $ \lambda_{inv\_fit} = \frac{1}{cl.fitness}$.  Figure~\ref{fig:gnmc-res} shows results of applying GNMC with these three mass functions to the XCSF instances trained on both environments, measured are the effect on performance ($\hat{Q}$ MAE and $\hat{\pi}$ accuracy) and population size (number of macro and microclassifiers).

In the deterministic case, GNMC retains performance when either $ \lambda_{fit} $ or $ \lambda_{tan} $ is used, for any value of $ \rho $. $ \hat{Q}$ MAE improves very slightly as $ \rho $ increases, and $\hat{\pi}$ accuracy is unchanged. Using $\lambda_{inv\_fit} $ gives smooth degradation in both metrics. Looking at the population sizes, both $\lambda_{fit} $ and $\lambda_{tan}$ exhibit a moderate reduction in the number of microclassifiers and a significant reduction in the number of macroclassifiers as $ \rho $ increases. However $\lambda_{inv\_fit}$ is different, yielding similar number of macroclassifiers in the extreme, but drastically smaller number of microclassifiers. This conforms to our expectations of its operation; classifiers that are kept by $\lambda_{inv\_fit} $ have low fitness values, and since low fitness classifiers tend to have low numerosity, the ratio of microclassifiers to macroclassifiers is low. For $ \lambda_{fit} $ and $\lambda_{tan}$ the ratio is much higher.
In the stochastic case, the situation is similar overall with a slight difference when considering the effect on performance: namely that $\hat{\pi}$ accuracy for both $\lambda_{fit}$ and $ \lambda_{tan}$ is degraded slightly instead of remaining constant as $ \rho $ increases.

\section{Rollout Analysis}
\label{sec:rollouts}
We now investigate the relationship between policy accuracy and the commonly used black-box performance metric, steps-to-goal (STG), from other works applying LCSs to maze-like RL environments, e.g. \cite{lanzi_extension_1999,lanzi_xcs_2005}. Note that in such environments, minimising STG is equivalent to maximising cumulative discounted reward. For our analysis we consider only the stochastic variant of FrozenLake8x8, as it showed the most interesting variations in $ \hat{\pi} $ accuracy when GNMC was applied to it in Section~\ref{sec:compaction}.

\begin{table}[t]
\centering
\scriptsize
\begin{tabular}{ll|l|l|l|l|l|l|l|l|l|l|l|l|l|l|}
\cline{3-6} \cline{8-11} \cline{13-16}
 &
   &
  \multicolumn{4}{c|}{\begin{tabular}[c]{@{}c@{}}No compaction\\ (Group A)\end{tabular}} &
  \textbf{} &
  \multicolumn{4}{c|}{\begin{tabular}[c]{@{}c@{}}GNMC $\lambda_{fit}\ \rho=0.99$\\ (Group B)\end{tabular}} &
  \textbf{} &
  \multicolumn{4}{c|}{\begin{tabular}[c]{@{}c@{}}GNMC $\lambda_{inv\_fit}\ \rho=0.99$\\ (Group C)\end{tabular}} \\ \cline{1-1} \cline{3-6} \cline{8-11} \cline{13-16} 
\multicolumn{1}{|c|}{\begin{tabular}[c]{@{}c@{}}Instance\\ num.\end{tabular}} &
  \multicolumn{1}{c|}{} &
  \multicolumn{1}{c|}{\cellcolor[HTML]{FFCCC9}\begin{tabular}[c]{@{}c@{}}Mean\\ STG\end{tabular}} &
  \multicolumn{1}{c|}{\cellcolor[HTML]{FFCE93}\begin{tabular}[c]{@{}c@{}}Max\\ STG\end{tabular}} &
  \multicolumn{1}{c|}{\cellcolor[HTML]{9AFF99}\begin{tabular}[c]{@{}c@{}}Num.\\ roll\end{tabular}} &
  \multicolumn{1}{c|}{\cellcolor[HTML]{96FFFB}\begin{tabular}[c]{@{}c@{}}$\hat{\pi}$\\ acc.\end{tabular}} &
  \multicolumn{1}{c|}{} &
  \multicolumn{1}{c|}{\cellcolor[HTML]{FFCCC9}\begin{tabular}[c]{@{}c@{}}Mean\\ STG\end{tabular}} &
  \multicolumn{1}{c|}{\cellcolor[HTML]{FFCE93}\begin{tabular}[c]{@{}c@{}}Max\\ STG\end{tabular}} &
  \multicolumn{1}{c|}{\cellcolor[HTML]{9AFF99}\begin{tabular}[c]{@{}c@{}}Num.\\ roll\end{tabular}} &
  \multicolumn{1}{c|}{\cellcolor[HTML]{96FFFB}\begin{tabular}[c]{@{}c@{}}$\hat{\pi}$\\ acc.\end{tabular}} &
  \multicolumn{1}{c|}{} &
  \multicolumn{1}{c|}{\cellcolor[HTML]{FFCCC9}\begin{tabular}[c]{@{}c@{}}Mean\\ STG\end{tabular}} &
  \multicolumn{1}{c|}{\cellcolor[HTML]{FFCE93}\begin{tabular}[c]{@{}c@{}}Max\\ STG\end{tabular}} &
  \multicolumn{1}{c|}{\cellcolor[HTML]{9AFF99}\begin{tabular}[c]{@{}c@{}}Num.\\ roll\end{tabular}} &
  \multicolumn{1}{c|}{\cellcolor[HTML]{96FFFB}\begin{tabular}[c]{@{}c@{}}$\hat{\pi}$\\ acc.\end{tabular}} \\ \cline{1-1} \cline{3-6} \cline{8-11} \cline{13-16}
  \\[-2ex]
  \cline{1-1} \cline{3-6} \cline{8-11} \cline{13-16}
\multicolumn{1}{|l|}{1} &
   &
  \cellcolor[HTML]{FFCCC9}15.42 &
  \cellcolor[HTML]{FFCE93}20 &
  \cellcolor[HTML]{9AFF99}114 &
  \cellcolor[HTML]{96FFFB}0.89 &
   &
  \cellcolor[HTML]{FFCCC9}15.31 &
  \cellcolor[HTML]{FFCE93}20 &
  \cellcolor[HTML]{9AFF99}117 &
  \cellcolor[HTML]{96FFFB}0.91 &
   &
  \cellcolor[HTML]{FFCCC9}15.44 &
  \cellcolor[HTML]{FFCE93}21 &
  \cellcolor[HTML]{9AFF99}114 &
  \cellcolor[HTML]{96FFFB}0.79 \\ \cline{1-1} \cline{3-6} \cline{8-11} \cline{13-16} 
\multicolumn{1}{|l|}{2} &
   &
  \cellcolor[HTML]{FFCCC9}15.39 &
  \cellcolor[HTML]{FFCE93}20 &
  \cellcolor[HTML]{9AFF99}116 &
  \cellcolor[HTML]{96FFFB}0.87 &
   &
  \cellcolor[HTML]{FFCCC9}15.43 &
  \cellcolor[HTML]{FFCE93}20 &
  \cellcolor[HTML]{9AFF99}117 &
  \cellcolor[HTML]{96FFFB}0.79 &
   &
  \cellcolor[HTML]{FFCCC9}15.45 &
  \cellcolor[HTML]{FFCE93}20 &
  \cellcolor[HTML]{9AFF99}115 &
  \cellcolor[HTML]{96FFFB}0.68 \\ \cline{1-1} \cline{3-6} \cline{8-11} \cline{13-16} 
\multicolumn{1}{|l|}{3} &
   &
  \cellcolor[HTML]{FFCCC9}32.36 &
  \cellcolor[HTML]{FFCE93}110 &
  \cellcolor[HTML]{9AFF99}123 &
  \cellcolor[HTML]{96FFFB}0.75 &
   &
  \cellcolor[HTML]{FFCCC9}* &
  \cellcolor[HTML]{FFCE93}* &
  \cellcolor[HTML]{9AFF99}\textbf{150} &
  \cellcolor[HTML]{96FFFB}0.72 &
   &
  \cellcolor[HTML]{FFCCC9}* &
  \cellcolor[HTML]{FFCE93}* &
  \cellcolor[HTML]{9AFF99}\textbf{150} &
  \cellcolor[HTML]{96FFFB}0.55 \\ \cline{1-1} \cline{3-6} \cline{8-11} \cline{13-16} 
\multicolumn{1}{|l|}{4} &
   &
  \cellcolor[HTML]{FFCCC9}15.37 &
  \cellcolor[HTML]{FFCE93}20 &
  \cellcolor[HTML]{9AFF99}127 &
  \cellcolor[HTML]{96FFFB}0.83 &
   &
  \cellcolor[HTML]{FFCCC9}15.46 &
  \cellcolor[HTML]{FFCE93}21 &
  \cellcolor[HTML]{9AFF99}127 &
  \cellcolor[HTML]{96FFFB}0.83 &
   &
  \cellcolor[HTML]{FFCCC9}18.74 &
  \cellcolor[HTML]{FFCE93}89 &
  \cellcolor[HTML]{9AFF99}131 &
  \cellcolor[HTML]{96FFFB}0.55 \\ \cline{1-1} \cline{3-6} \cline{8-11} \cline{13-16} 
\multicolumn{1}{|l|}{5} &
   &
  \cellcolor[HTML]{FFCCC9}15.38 &
  \cellcolor[HTML]{FFCE93}20 &
  \cellcolor[HTML]{9AFF99}116 &
  \cellcolor[HTML]{96FFFB}0.83 &
   &
  \cellcolor[HTML]{FFCCC9}15.38 &
  \cellcolor[HTML]{FFCE93}20 &
  \cellcolor[HTML]{9AFF99}116 &
  \cellcolor[HTML]{96FFFB}0.85 &
   &
  \cellcolor[HTML]{FFCCC9}15.38 &
  \cellcolor[HTML]{FFCE93}20 &
  \cellcolor[HTML]{9AFF99}116 &
  \cellcolor[HTML]{96FFFB}0.70 \\ \cline{1-1} \cline{3-6} \cline{8-11} \cline{13-16} 
\multicolumn{1}{|l|}{6} &
   &
  \cellcolor[HTML]{FFCCC9}15.38 &
  \cellcolor[HTML]{FFCE93}20 &
  \cellcolor[HTML]{9AFF99}116 &
  \cellcolor[HTML]{96FFFB}0.85 &
   &
  \cellcolor[HTML]{FFCCC9}16.13 &
  \cellcolor[HTML]{FFCE93}25 &
  \cellcolor[HTML]{9AFF99}126 &
  \cellcolor[HTML]{96FFFB}0.79 &
   &
  \cellcolor[HTML]{FFCCC9}33.11 &
  \cellcolor[HTML]{FFCE93}148 &
  \cellcolor[HTML]{9AFF99}115 &
  \cellcolor[HTML]{96FFFB}0.66 \\ \cline{1-1} \cline{3-6} \cline{8-11} \cline{13-16} 
\multicolumn{1}{|l|}{7} &
   &
  \cellcolor[HTML]{FFCCC9}15.38 &
  \cellcolor[HTML]{FFCE93}20 &
  \cellcolor[HTML]{9AFF99}116 &
  \cellcolor[HTML]{96FFFB}0.87 &
   &
  \cellcolor[HTML]{FFCCC9}15.38 &
  \cellcolor[HTML]{FFCE93}20 &
  \cellcolor[HTML]{9AFF99}116 &
  \cellcolor[HTML]{96FFFB}0.85 &
   &
  \cellcolor[HTML]{FFCCC9}15.38 &
  \cellcolor[HTML]{FFCE93}20 &
  \cellcolor[HTML]{9AFF99}116 &
  \cellcolor[HTML]{96FFFB}0.74 \\ \cline{1-1} \cline{3-6} \cline{8-11} \cline{13-16} 
\multicolumn{1}{|l|}{8} &
   &
  \cellcolor[HTML]{FFCCC9}15.38 &
  \cellcolor[HTML]{FFCE93}20 &
  \cellcolor[HTML]{9AFF99}116 &
  \cellcolor[HTML]{96FFFB}0.87 &
   &
  \cellcolor[HTML]{FFCCC9}15.38 &
  \cellcolor[HTML]{FFCE93}20 &
  \cellcolor[HTML]{9AFF99}116 &
  \cellcolor[HTML]{96FFFB}0.85 &
   &
  \cellcolor[HTML]{FFCCC9}37.46 &
  \cellcolor[HTML]{FFCE93}98 &
  \cellcolor[HTML]{9AFF99}142 &
  \cellcolor[HTML]{96FFFB}0.60 \\ \cline{1-1} \cline{3-6} \cline{8-11} \cline{13-16} 
\multicolumn{1}{|l|}{9} &
   &
  \cellcolor[HTML]{FFCCC9}15.43 &
  \cellcolor[HTML]{FFCE93}20 &
  \cellcolor[HTML]{9AFF99}115 &
  \cellcolor[HTML]{96FFFB}0.83 &
   &
  \cellcolor[HTML]{FFCCC9}33.25 &
  \cellcolor[HTML]{FFCE93}93 &
  \cellcolor[HTML]{9AFF99}118 &
  \cellcolor[HTML]{96FFFB}0.81 &
   &
  \cellcolor[HTML]{FFCCC9}15.39 &
  \cellcolor[HTML]{FFCE93}20 &
  \cellcolor[HTML]{9AFF99}116 &
  \cellcolor[HTML]{96FFFB}0.75 \\ \cline{1-1} \cline{3-6} \cline{8-11} \cline{13-16} 
\multicolumn{1}{|l|}{10} &
   &
  \cellcolor[HTML]{FFCCC9}15.31 &
  \cellcolor[HTML]{FFCE93}20 &
  \cellcolor[HTML]{9AFF99}117 &
  \cellcolor[HTML]{96FFFB}0.85 &
   &
  \cellcolor[HTML]{FFCCC9}15.31 &
  \cellcolor[HTML]{FFCE93}20 &
  \cellcolor[HTML]{9AFF99}117 &
  \cellcolor[HTML]{96FFFB}0.91 &
   &
  \cellcolor[HTML]{FFCCC9}15.70 &
  \cellcolor[HTML]{FFCE93}23 &
  \cellcolor[HTML]{9AFF99}117 &
  \cellcolor[HTML]{96FFFB}0.68 \\ \cline{1-1} \cline{3-6} \cline{8-11} \cline{13-16} 
\multicolumn{1}{|l|}{11} &
   &
  \cellcolor[HTML]{FFCCC9}15.38 &
  \cellcolor[HTML]{FFCE93}20 &
  \cellcolor[HTML]{9AFF99}116 &
  \cellcolor[HTML]{96FFFB}0.87 &
   &
  \cellcolor[HTML]{FFCCC9}15.45 &
  \cellcolor[HTML]{FFCE93}20 &
  \cellcolor[HTML]{9AFF99}115 &
  \cellcolor[HTML]{96FFFB}0.89 &
   &
  \cellcolor[HTML]{FFCCC9}* &
  \cellcolor[HTML]{FFCE93}* &
  \cellcolor[HTML]{9AFF99}\textbf{150} &
  \cellcolor[HTML]{96FFFB}0.55 \\ \cline{1-1} \cline{3-6} \cline{8-11} \cline{13-16} 
\multicolumn{1}{|l|}{12} &
   &
  \cellcolor[HTML]{FFCCC9}15.66 &
  \cellcolor[HTML]{FFCE93}20 &
  \cellcolor[HTML]{9AFF99}113 &
  \cellcolor[HTML]{96FFFB}0.77 &
   &
  \cellcolor[HTML]{FFCCC9}15.39 &
  \cellcolor[HTML]{FFCE93}20 &
  \cellcolor[HTML]{9AFF99}116 &
  \cellcolor[HTML]{96FFFB}0.77 &
   &
  \cellcolor[HTML]{FFCCC9}* &
  \cellcolor[HTML]{FFCE93}* &
  \cellcolor[HTML]{9AFF99}\textbf{150} &
  \cellcolor[HTML]{96FFFB}0.64 \\ \cline{1-1} \cline{3-6} \cline{8-11} \cline{13-16} 
\multicolumn{1}{|l|}{13} &
   &
  \cellcolor[HTML]{FFCCC9}37.61 &
  \cellcolor[HTML]{FFCE93}119 &
  \cellcolor[HTML]{9AFF99}130 &
  \cellcolor[HTML]{96FFFB}0.75 &
   &
  \cellcolor[HTML]{FFCCC9}37.43 &
  \cellcolor[HTML]{FFCE93}119 &
  \cellcolor[HTML]{9AFF99}128 &
  \cellcolor[HTML]{96FFFB}0.75 &
   &
  \cellcolor[HTML]{FFCCC9}* &
  \cellcolor[HTML]{FFCE93}* &
  \cellcolor[HTML]{9AFF99}\textbf{150} &
  \cellcolor[HTML]{96FFFB}0.57 \\ \cline{1-1} \cline{3-6} \cline{8-11} \cline{13-16} 
\multicolumn{1}{|l|}{14} &
   &
  \cellcolor[HTML]{FFCCC9}15.38 &
  \cellcolor[HTML]{FFCE93}20 &
  \cellcolor[HTML]{9AFF99}116 &
  \cellcolor[HTML]{96FFFB}0.83 &
   &
  \cellcolor[HTML]{FFCCC9}15.33 &
  \cellcolor[HTML]{FFCE93}20 &
  \cellcolor[HTML]{9AFF99}117 &
  \cellcolor[HTML]{96FFFB}0.79 &
   &
  \cellcolor[HTML]{FFCCC9}47.94 &
  \cellcolor[HTML]{FFCE93}143 &
  \cellcolor[HTML]{9AFF99}119 &
  \cellcolor[HTML]{96FFFB}0.64 \\ \cline{1-1} \cline{3-6} \cline{8-11} \cline{13-16} 
\multicolumn{1}{|l|}{15} &
   &
  \cellcolor[HTML]{FFCCC9}15.45 &
  \cellcolor[HTML]{FFCE93}20 &
  \cellcolor[HTML]{9AFF99}115 &
  \cellcolor[HTML]{96FFFB}0.85 &
   &
  \cellcolor[HTML]{FFCCC9}15.43 &
  \cellcolor[HTML]{FFCE93}20 &
  \cellcolor[HTML]{9AFF99}117 &
  \cellcolor[HTML]{96FFFB}0.83 &
   &
  \cellcolor[HTML]{FFCCC9}15.44 &
  \cellcolor[HTML]{FFCE93}21 &
  \cellcolor[HTML]{9AFF99}114 &
  \cellcolor[HTML]{96FFFB}0.68 \\ \cline{1-1} \cline{3-6} \cline{8-11} \cline{13-16} 
\multicolumn{1}{|l|}{16} &
   &
  \cellcolor[HTML]{FFCCC9}15.38 &
  \cellcolor[HTML]{FFCE93}20 &
  \cellcolor[HTML]{9AFF99}116 &
  \cellcolor[HTML]{96FFFB}0.91 &
   &
  \cellcolor[HTML]{FFCCC9}15.42 &
  \cellcolor[HTML]{FFCE93}20 &
  \cellcolor[HTML]{9AFF99}114 &
  \cellcolor[HTML]{96FFFB}0.92 &
   &
  \cellcolor[HTML]{FFCCC9}37.22 &
  \cellcolor[HTML]{FFCE93}112 &
  \cellcolor[HTML]{9AFF99}120 &
  \cellcolor[HTML]{96FFFB}0.55 \\ \cline{1-1} \cline{3-6} \cline{8-11} \cline{13-16} 
\multicolumn{1}{|l|}{17} &
   &
  \cellcolor[HTML]{FFCCC9}15.43 &
  \cellcolor[HTML]{FFCE93}21 &
  \cellcolor[HTML]{9AFF99}122 &
  \cellcolor[HTML]{96FFFB}0.83 &
   &
  \cellcolor[HTML]{FFCCC9}15.32 &
  \cellcolor[HTML]{FFCE93}21 &
  \cellcolor[HTML]{9AFF99}116 &
  \cellcolor[HTML]{96FFFB}0.83 &
   &
  \cellcolor[HTML]{FFCCC9}15.48 &
  \cellcolor[HTML]{FFCE93}21 &
  \cellcolor[HTML]{9AFF99}129 &
  \cellcolor[HTML]{96FFFB}0.64 \\ \cline{1-1} \cline{3-6} \cline{8-11} \cline{13-16} 
\multicolumn{1}{|l|}{18} &
   &
  \cellcolor[HTML]{FFCCC9}15.38 &
  \cellcolor[HTML]{FFCE93}20 &
  \cellcolor[HTML]{9AFF99}116 &
  \cellcolor[HTML]{96FFFB}0.89 &
   &
  \cellcolor[HTML]{FFCCC9}15.37 &
  \cellcolor[HTML]{FFCE93}20 &
  \cellcolor[HTML]{9AFF99}119 &
  \cellcolor[HTML]{96FFFB}0.89 &
   &
  \cellcolor[HTML]{FFCCC9}* &
  \cellcolor[HTML]{FFCE93}* &
  \cellcolor[HTML]{9AFF99}\textbf{150} &
  \cellcolor[HTML]{96FFFB}0.72 \\ \cline{1-1} \cline{3-6} \cline{8-11} \cline{13-16} 
\multicolumn{1}{|l|}{19} &
   &
  \cellcolor[HTML]{FFCCC9}15.38 &
  \cellcolor[HTML]{FFCE93}20 &
  \cellcolor[HTML]{9AFF99}116 &
  \cellcolor[HTML]{96FFFB}0.85 &
   &
  \cellcolor[HTML]{FFCCC9}15.38 &
  \cellcolor[HTML]{FFCE93}20 &
  \cellcolor[HTML]{9AFF99}116 &
  \cellcolor[HTML]{96FFFB}0.81 &
   &
  \cellcolor[HTML]{FFCCC9}15.27 &
  \cellcolor[HTML]{FFCE93}19 &
  \cellcolor[HTML]{9AFF99}121 &
  \cellcolor[HTML]{96FFFB}0.72 \\ \cline{1-1} \cline{3-6} \cline{8-11} \cline{13-16} 
\multicolumn{1}{|l|}{20} &
   &
  \cellcolor[HTML]{FFCCC9}15.92 &
  \cellcolor[HTML]{FFCE93}21 &
  \cellcolor[HTML]{9AFF99}118 &
  \cellcolor[HTML]{96FFFB}0.79 &
   &
  \cellcolor[HTML]{FFCCC9}16.06 &
  \cellcolor[HTML]{FFCE93}25 &
  \cellcolor[HTML]{9AFF99}128 &
  \cellcolor[HTML]{96FFFB}0.75 &
   &
  \cellcolor[HTML]{FFCCC9}* &
  \cellcolor[HTML]{FFCE93}* &
  \cellcolor[HTML]{9AFF99}\textbf{150} &
  \cellcolor[HTML]{96FFFB}0.58 \\ \cline{1-1} \cline{3-6} \cline{8-11} \cline{13-16} 
\multicolumn{1}{|l|}{21} &
   &
  \cellcolor[HTML]{FFCCC9}\textbf{67.29} &
  \cellcolor[HTML]{FFCE93}\textbf{193} &
  \cellcolor[HTML]{9AFF99}110 &
  \cellcolor[HTML]{96FFFB}0.79 &
   &
  \cellcolor[HTML]{FFCCC9}\textbf{68.61} &
  \cellcolor[HTML]{FFCE93}\textbf{193} &
  \cellcolor[HTML]{9AFF99}117 &
  \cellcolor[HTML]{96FFFB}0.75 &
   &
  \cellcolor[HTML]{FFCCC9}53.88 &
  \cellcolor[HTML]{FFCE93}110 &
  \cellcolor[HTML]{9AFF99}121 &
  \cellcolor[HTML]{96FFFB}0.57 \\ \cline{1-1} \cline{3-6} \cline{8-11} \cline{13-16} 
\multicolumn{1}{|l|}{22} &
   &
  \cellcolor[HTML]{FFCCC9}15.38 &
  \cellcolor[HTML]{FFCE93}20 &
  \cellcolor[HTML]{9AFF99}116 &
  \cellcolor[HTML]{96FFFB}0.85 &
   &
  \cellcolor[HTML]{FFCCC9}32.79 &
  \cellcolor[HTML]{FFCE93}110 &
  \cellcolor[HTML]{9AFF99}123 &
  \cellcolor[HTML]{96FFFB}0.79 &
   &
  \cellcolor[HTML]{FFCCC9}62.20 &
  \cellcolor[HTML]{FFCE93}\textbf{200} &
  \cellcolor[HTML]{9AFF99}124 &
  \cellcolor[HTML]{96FFFB}0.70 \\ \cline{1-1} \cline{3-6} \cline{8-11} \cline{13-16} 
\multicolumn{1}{|l|}{23} &
   &
  \cellcolor[HTML]{FFCCC9}15.38 &
  \cellcolor[HTML]{FFCE93}20 &
  \cellcolor[HTML]{9AFF99}116 &
  \cellcolor[HTML]{96FFFB}0.85 &
   &
  \cellcolor[HTML]{FFCCC9}15.33 &
  \cellcolor[HTML]{FFCE93}20 &
  \cellcolor[HTML]{9AFF99}117 &
  \cellcolor[HTML]{96FFFB}0.79 &
   &
  \cellcolor[HTML]{FFCCC9}53.57 &
  \cellcolor[HTML]{FFCE93}153 &
  \cellcolor[HTML]{9AFF99}115 &
  \cellcolor[HTML]{96FFFB}0.53 \\ \cline{1-1} \cline{3-6} \cline{8-11} \cline{13-16} 
\multicolumn{1}{|l|}{24} &
   &
  \cellcolor[HTML]{FFCCC9}32.91 &
  \cellcolor[HTML]{FFCE93}100 &
  \cellcolor[HTML]{9AFF99}112 &
  \cellcolor[HTML]{96FFFB}0.79 &
   &
  \cellcolor[HTML]{FFCCC9}15.33 &
  \cellcolor[HTML]{FFCE93}20 &
  \cellcolor[HTML]{9AFF99}117 &
  \cellcolor[HTML]{96FFFB}0.75 &
   &
  \cellcolor[HTML]{FFCCC9}* &
  \cellcolor[HTML]{FFCE93}* &
  \cellcolor[HTML]{9AFF99}\textbf{150} &
  \cellcolor[HTML]{96FFFB}0.60 \\ \cline{1-1} \cline{3-6} \cline{8-11} \cline{13-16} 
\multicolumn{1}{|l|}{25} &
   &
  \cellcolor[HTML]{FFCCC9}15.38 &
  \cellcolor[HTML]{FFCE93}20 &
  \cellcolor[HTML]{9AFF99}116 &
  \cellcolor[HTML]{96FFFB}0.89 &
   &
  \cellcolor[HTML]{FFCCC9}15.38 &
  \cellcolor[HTML]{FFCE93}20 &
  \cellcolor[HTML]{9AFF99}116 &
  \cellcolor[HTML]{96FFFB}0.87 &
   &
  \cellcolor[HTML]{FFCCC9}28.51 &
  \cellcolor[HTML]{FFCE93}86 &
  \cellcolor[HTML]{9AFF99}119 &
  \cellcolor[HTML]{96FFFB}0.62 \\ \cline{1-1} \cline{3-6} \cline{8-11} \cline{13-16} 
\multicolumn{1}{|l|}{26} &
   &
  \cellcolor[HTML]{FFCCC9}17.21 &
  \cellcolor[HTML]{FFCE93}82 &
  \cellcolor[HTML]{9AFF99}119 &
  \cellcolor[HTML]{96FFFB}0.81 &
   &
  \cellcolor[HTML]{FFCCC9}15.85 &
  \cellcolor[HTML]{FFCE93}25 &
  \cellcolor[HTML]{9AFF99}138 &
  \cellcolor[HTML]{96FFFB}0.87 &
   &
  \cellcolor[HTML]{FFCCC9}* &
  \cellcolor[HTML]{FFCE93}* &
  \cellcolor[HTML]{9AFF99}\textbf{150} &
  \cellcolor[HTML]{96FFFB}\textbf{0.38} \\ \cline{1-1} \cline{3-6} \cline{8-11} \cline{13-16} 
\multicolumn{1}{|l|}{27} &
   &
  \cellcolor[HTML]{FFCCC9}15.74 &
  \cellcolor[HTML]{FFCE93}21 &
  \cellcolor[HTML]{9AFF99}\textbf{142} &
  \cellcolor[HTML]{96FFFB}\textbf{0.74} &
   &
  \cellcolor[HTML]{FFCCC9}15.96 &
  \cellcolor[HTML]{FFCE93}40 &
  \cellcolor[HTML]{9AFF99}133 &
  \cellcolor[HTML]{96FFFB}0.72 &
   &
  \cellcolor[HTML]{FFCCC9}* &
  \cellcolor[HTML]{FFCE93}* &
  \cellcolor[HTML]{9AFF99}\textbf{150} &
  \cellcolor[HTML]{96FFFB}0.47 \\ \cline{1-1} \cline{3-6} \cline{8-11} \cline{13-16} 
\multicolumn{1}{|l|}{28} &
   &
  \cellcolor[HTML]{FFCCC9}36.96 &
  \cellcolor[HTML]{FFCE93}139 &
  \cellcolor[HTML]{9AFF99}126 &
  \cellcolor[HTML]{96FFFB}0.79 &
   &
  \cellcolor[HTML]{FFCCC9}* &
  \cellcolor[HTML]{FFCE93}* &
  \cellcolor[HTML]{9AFF99}\textbf{150} &
  \cellcolor[HTML]{96FFFB}\textbf{0.70} &
   &
  \cellcolor[HTML]{FFCCC9}* &
  \cellcolor[HTML]{FFCE93}* &
  \cellcolor[HTML]{9AFF99}\textbf{150} &
  \cellcolor[HTML]{96FFFB}0.47 \\ \cline{1-1} \cline{3-6} \cline{8-11} \cline{13-16} 
\multicolumn{1}{|l|}{29} &
   &
  \cellcolor[HTML]{FFCCC9}15.38 &
  \cellcolor[HTML]{FFCE93}20 &
  \cellcolor[HTML]{9AFF99}116 &
  \cellcolor[HTML]{96FFFB}0.83 &
   &
  \cellcolor[HTML]{FFCCC9}15.37 &
  \cellcolor[HTML]{FFCE93}19 &
  \cellcolor[HTML]{9AFF99}119 &
  \cellcolor[HTML]{96FFFB}0.77 &
   &
  \cellcolor[HTML]{FFCCC9}\textbf{82.36} &
  \cellcolor[HTML]{FFCE93}197 &
  \cellcolor[HTML]{9AFF99}114 &
  \cellcolor[HTML]{96FFFB}0.62 \\ \cline{1-1} \cline{3-6} \cline{8-11} \cline{13-16} 
\multicolumn{1}{|l|}{30} &
   &
  \cellcolor[HTML]{FFCCC9}15.38 &
  \cellcolor[HTML]{FFCE93}20 &
  \cellcolor[HTML]{9AFF99}116 &
  \cellcolor[HTML]{96FFFB}0.83 &
   &
  \cellcolor[HTML]{FFCCC9}15.38 &
  \cellcolor[HTML]{FFCE93}20 &
  \cellcolor[HTML]{9AFF99}116 &
  \cellcolor[HTML]{96FFFB}0.81 &
   &
  \cellcolor[HTML]{FFCCC9}* &
  \cellcolor[HTML]{FFCE93}* &
  \cellcolor[HTML]{9AFF99}\textbf{150} &
  \cellcolor[HTML]{96FFFB}0.55 \\ \cline{1-1} \cline{3-6} \cline{8-11} \cline{13-16} 
\end{tabular}
\vspace{.5em}
\caption{Results of STG testing procedure on FrozenLake8x8 $ p_{slip} = 0.1 $ for three different groups of XCSF instances. Asterisks for mean and max STG indicate incomplete data. Set in bold are the ``worst'' values for each column ($\hat{\pi}$ acc. minimum, others maximum).}
\label{tab:rollout-data}
\end{table}
A testing procedure to measure STG is devised as follows: allow each XCSF instance a budget of 150 rollouts (episodes) and in these 150 rollouts, attempt to record STG (successfully reach the goal) 100 times. If 100 successes are not achieved, STG data is incomplete. In each rollout, the agent's initial state is $ (0, 0) $ and the random seed of the environment is set to a unique number. Note this is different from training where the agent's initial state was any $ s \in S $, selected uniformly at random. This testing procedure is applied to all 30 trained XCSF instances in three groups, representing different levels of GNMC compaction: no compaction, compaction with $\lambda_{fit}$ $\rho=0.99$,
and compaction with $\lambda_{inv\_fit}$ $\rho=0.99$. Table~\ref{tab:rollout-data} shows the collected results: included are mean and max STG, number of rollouts performed, and $\hat{\pi}$ accuracy (for reference). Note that minimum STG in the environment is 14.

Group A in general achieves admirable mean STG; only 5 out of 30 instances could be considered as outliers. Number of rollouts for all instances is generally not much higher than 110. This indicates that in most instances XCSF is quickly navigating towards the goal, with a low failure rate due to environmental stochasticity. Comparing Group B to Group A, there are two instances in Group B that have failed to collect complete STG data, also having the two lowest policy accuracies. In general, all four measures degrade only slightly between the two groups, which indicates that the compaction applied to Group B is not having a detrimental effect on performance. Transitioning from Group A to Group C however produces noticeable performance loss. 11 out of 30 instances fail to record complete STG data, and generally those that do show degradation in all four measures.

It is difficult to determine the exact relationship between STG and policy accuracy. Some instances (e.g. instance 1) exhibit minimal degradation in both measures between groups, but some exhibit larger degradation (e.g. instance 22). In cases where degradation in policy accuracy is small but degradation in STG is large, the cause is often a minority of states along the edge of the grid that are advocating actions where the only possible way to advance further towards the goal is to slip, e.g. advocating $ Right $ in any of the states in the rightmost column (where $ x = 7 $).
It is therefore clear that the two metrics are complementary rather than competitive. Policy accuracy is measured globally and is always defined, and STG is measured on a \textit{specific task} (starting state), possibly being undefined/incomplete. The starting state is an arbitrary choice and if altered results in differing STG but unchanged policy accuracy.
\section{Conclusion}
\label{sec:conclusion}
We trained XCSF on a deterministic and stochastic variant of FrozenLake8x8, measuring its performance with respect to the optimal solutions produced via dynamic programming. Results show that in both cases XCSF achieved low Q-function approximation error, and in the deterministic case XCSF converged to maximum policy accuracy. In the stochastic case policy accuracy was noticeably degraded both because of increased problem difficulty and increased strictness of the optimal policy. Next we introduced Greedy Niche Mass Compaction (GNMC), a compaction algorithm designed for LCSs applied to discrete RL environments. We showed GNMC is a generalisation of previous work and applied it to our trained XCSF instances. Given a suitable mass function, GNMC can yield a significant reduction in population size without increasing function approximation error and only slightly decreasing policy accuracy. Finally we linked our policy accuracy metric to the steps-to-goal metric used in previous work across multiple groups of compacted XCSF instances. This highlighted how the two metrics are complementary rather than competitive. Suggested future work includes applying GNMC to environments where populations are larger/more complex and the mass removal factor has more impact on performance. GNMC's concept could also be extended to continuous state and/or action spaces.

%
%
\bibliographystyle{splncs04}
\bibliography{refs.bib}

\end{document}